\def\year{2021}\relax
\documentclass[letterpaper]{article} 
\usepackage{aaai21}  
\usepackage{times}  
\usepackage{helvet} 
\usepackage{courier}  
\usepackage[hyphens]{url}  
\usepackage{graphicx} 
\usepackage{graphicx}  
\usepackage{natbib}  
\usepackage{caption} 
\usepackage[normalem]{ulem}
\usepackage{enumitem,setspace}
\usepackage{amsmath,amssymb,amsthm}
\usepackage{multirow}
\usepackage{makecell}
\usepackage{arydshln}
\usepackage{color}
\usepackage{booktabs}
\usepackage[switch]{lineno}
\usepackage{algorithm}
\usepackage{algorithmic}

\title{Enhancing Scientific Papers Summarization with Citation Graph}

\author{Chenxin An,
Ming Zhong,
Yiran Chen,
Danqing Wang,
Xipeng Qiu\thanks{\ \  Corresponding author.},
Xuanjing Huang\\
}
\affiliations{
    {Shanghai Key Laboratory of Intelligent Information Processing, Fudan University\\
    School of Computer Science, Fudan University}\\
    \{cxan20, mzhong18, yrchen19, dqwang18, xpqiu, xjhuang\}@fudan.edu.cn
}

\begin{document}
\maketitle

\begin{abstract}
Previous work for text summarization in scientific domain mainly focused on the content of the input document, but seldom considering its citation network.
However, scientific papers are full of uncommon domain-specific terms, making it almost impossible for the model to understand its true meaning without the help of the relevant research community.
In this paper, we redefine the task of scientific papers summarization by utilizing their citation graph and propose a citation graph-based summarization model (\textsc{CgSum}) which can incorporate the information of both the source paper and its references.
In addition,  we construct a novel scientific papers summarization dataset Semantic Scholar Network (SSN) which contains 141K research papers in different domains and 661K citation relationships. The entire dataset constitutes a large connected citation graph.
Extensive experiments show that our model can achieve competitive performance when compared with the pretrained models even with a simple architecture.
The results also indicates the citation graph is crucial to better understand the content of papers and generate high-quality summaries.
Our code and dataset are available on  {\url{https://github.com/ChenxinAn-fdu/CGSum}}
\end{abstract}

\section{Introduction}
\begin{figure}[t]
\centering
\includegraphics[width=0.48\textwidth]{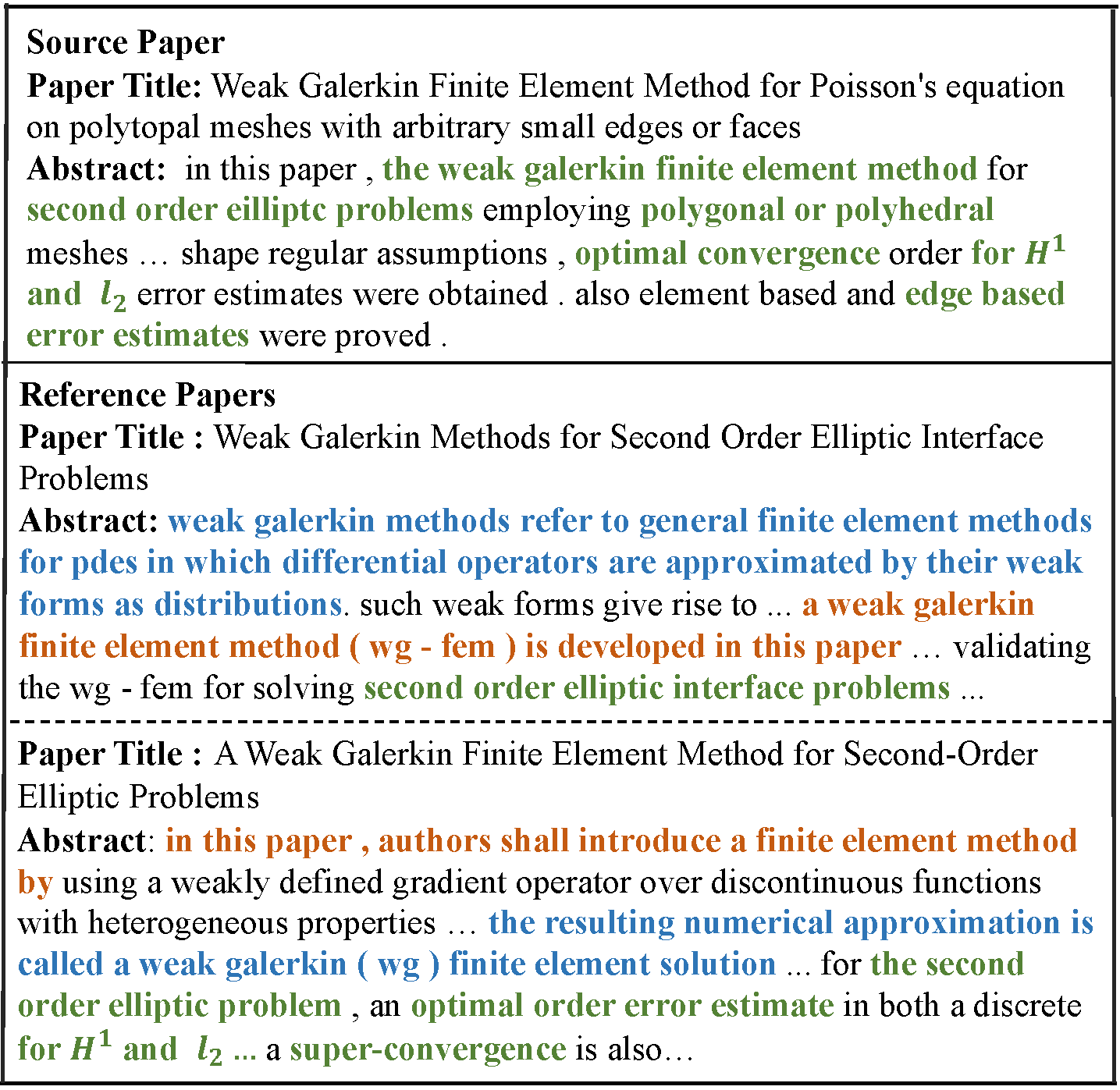}
\caption{A small research community on the subject of \textit{Weak Galerkin Finite Element Method}. Green text indicates the domain-specific terms shared in these papers, orange text denotes different ways of writing the same sentences, blue text represents the definition of \textit{Weak Galerkin Finite Element Method} (does not appear in the source paper).
}
\label{fig:example}
\end{figure}

Text summarization is to automatically compress a document into a shorter version preserving a concise description of the content. Most of the previous work focused on News domain \cite{nallapati2016abstractive,rush2015neural,nallapati2016summarunner,zhong2019searching}, and achieved promising result using the neural encoder-decoder architecture. Although text summarization systems have not been explored too much in other domains, such as scientific papers, they still have broad application prospects.

Generating a good abstract for a scientific paper is a very challenging task, even for a beginner researcher, since the scientific papers are usually longer and full of complex concepts and domain-specific items in specific fields. \citet{cohan2018discourse} and \citet{xiao2019extractive} leveraged the paper structure information to generate the abstracts for scientific papers. However, their methods dedicate to solving the problem of long document modeling and do not utilize the information of references. As a matter of fact, researchers usually write an abstract of a paper by referring some examples. Especially a large number of papers on the same topic are often similar in content.  Reasonable use of the information of reference papers may help us solve the scientific papers summarization task.
To generate better summary for a scientific paper, \citet{yasunaga2019scisummnet} integrated the formation of the source paper and the papers which cite the source papers. However, the citing papers appeared after the source paper, so we tend to think that this task does not help a research to draft an abstract when the paper has not been cited yet. 

In this paper, we highlight the importance of the citation graph and believe that it can assist in generating high-quality summaries. Figure \ref{fig:example} shows an example of a small research community consisting of the source paper and several reference papers. They are all about topic \textit{Weak Galerkin Finite Element Method} and are thus very similar in content, logic, and writing style. For instance, many uncommon domain-specific terms (green text) are shared in these papers, it is almost impossible for the model to understand the true meaning of these concepts without sufficient descriptions, so naturally, we should encourage the models to learn from the reference papers. The same expression always has different writing styles (orange text) in different papers, even some academic definitions that do not appear in the original text can be found in other papers (blue text), this relevant information will undoubtedly help the model to better understand the entire research community.

Motivated by the above observations, we augment the task of scientific papers summarization with citation graph. While generating the abstract of a source paper, the summarization systems are able to refer to papers in the same research community. Considering that all current large-scale scientific summarization datasets do not provide citation relationships between papers, we construct a scientific papers summarization dataset \textit{Semantic Scholar Network (SSN)} which contains 141K papers and 661K citation relationships extracted from the Semantic Scholar Open Research Corpus (S2ORC) \cite{lo2020s2orc}. Notably, our dataset is a huge connected citation graph, and each paper has class labels denoting its research field.
We divide the enhanced summarization task into 2 settings: (1) \textbf{transductive}: during training, models can access to all the nodes and edges in the whole dataset including papers (excluding abstracts) in the test set. (2) \textbf{inductive}: papers in the test set are from a totally new graph which means all test nodes cannot be used during training. 

Further, we propose a citation graph-based summarization model (\textsc{CgSum}). which incorporates the document and relevant citation graph when generating summaries. For each source paper, we obtain its corresponding research community by sampling a subgraph from the whole citation graph. We firstly encode the content of the source paper and utilize a graph encoder to capture the information of the subgraph. Finally, a decoder combines all the features outputted by the two encoders to produce the final summary. Additionally, we introduce a novel ROUGE credit method, which can instruct the model how to write summaries with the help of other papers' abstract in the same research community. Although our model only uses BiLSTM and GNN structures, experimental results show that  it achieves the competitive performance when compared with the pretrained model. We summarize our contributions as follows:
\begin{itemize}
\item We augment the task of scientific papers summarization by introducing the citation graph. 

\item We construct a large-scale summarization dataset SSN. To our best knowledge, this is the first large-scale scientific papers summarization dataset with citation graph.
\item We propose a citation graph-based summarization model to solve the enhanced task of scientific papers summarization, which can incorporate the source paper information and the features of the citation graph at the same time. 
\end{itemize}

\begin{table*}[!htbp]
    \begin{small}
        \centerline{
            \begin{tabular}{llcccccccc}
                \toprule
                \multicolumn{1}{l}{\multirow{2}[1]{*}{\textbf{Datasets}}} &
                \multicolumn{1}{l}{\multirow{2}[1]{*}{\textbf{Source}}} &
                \multicolumn{3}{c}{\textbf{\# Pairs}} & 
                 \multicolumn{2}{c}{\textbf{Doc. Length}}
                & \multicolumn{2}{c}{\textbf{Sum. Length}} & 
                \multicolumn{1}{c}{\multirow{2}[1]{*}{\textbf{\# Sections}}}
                \\
                & & Train & Val & Test & \# Words & \# Sent. & \# Words &
                \# Sent. & 
                \\ 
                \midrule
                CNN           & News& 90,266&1,220&1,093                                           & 760.5                    & 34.0                          & 45.7                     & 3.6   & -                                                    \\
                DailyMail      & News        & 196,961&12,148&10,397                                          & 653.3                    & 29.3                          & 54.7                     & 3.9   & -                                                      \\
                ScisummNet             & Scientific Papers & 1009 &-- &-- 
                & 4203.4 & 178.0 & 150.7 & 7.4 & 6.5\\
                arXiv$^\dag$          &Scientific Papers  & 215,913 &6440&6436                                           & 4938.0                    & 206.3                          & 220.0                    & 9.6  & 5.9                                                        \\
                PubMed$^\dag$       &Scientific Papers  & 119,924 &6633& 6658                                              & 3016.0                    & 86.4                          & 203.0                     & 6.9        & 5.6                                          \\
                SSN (inductive)   &\multirow{2}*{Scientific Papers} & 128,400 &
                6123& 6276 & \multirow{2}*{5072.3}           & \multirow{2}*{290.6}   & \multirow{2}*{165.1}      & \multirow{2}*{6.4}  & \multirow{2}*{10.8}\\
                SSN (transductive)&   & 128,299 & 6250& 6250 \\
                \bottomrule
            \end{tabular}
        }
    \end{small}
    
    \caption{\label{tab:statistics}
    Dataset statistics. The datasets with $^\dag$ indicates that the reported data comes from \citet{cohan2018discourse}. 
    }  
\end{table*}

\section{Related Work}

\subsection{Summarization with Graph Structures}
Early approaches for extractive summarization, such as TextRank \cite{mihalcea2004textrank}, have taken advantage of graph structures by building the connectivity graph with inter-sentence cosine similarity. As for the neural systems ,\citet{wang2020heterogeneous}  construct a heterogeneous graph network to model the relations between different semantic units.
On abstractive system, inspired by the great success of Graph Attention Networks (GATs) \cite{velivckovic2017graph} in NLP, \citet{song-etal-2018-graph} proposed the task of text generation from graph and  \citet{koncel-kedziorski-etal-2019-text} design a GATs-based transformer encoder to generate summary with the help of knowledge graphs extracted from scientific texts. For the combination of text and graph, \citet{fernandes2018structured} incorporates the regular document encoder with graph neural networks to make use of both the input sequence and graph structure, and \citet{zhu2020boosting,huang2020knowledge} built a knowledge graph from the input document and integrated it into the decoding process. Instead of directly generating abstract from the graph, our model uses the graph-enhanced encoder, viewing the citation graph as complementary information.

\subsection{Scientific Papers Summarization}
Automatic summarization for scientific papers has been studied for decades. Previous work mainly focused on the content of document \cite{luhn1958automatic, cohan2018scientific} and most of them are extractive \cite{teufel2002summarizing, xiao2019extractive}. For instance, \citet{cohan2018discourse} propose a neural model under the sequence-to-sequence framework with the discourse structure of scientific papers. These methods focus on modeling long documents, but ignore the influence of the research community it belongs to.
Another direction is citation summarization \cite{qazvinian2008scientific, cohan2018scientific,yasunaga2019scisummnet}, which can make use of the reference relationship between papers. Citation summarization aims to generate the summary of a source Paper according to the papers citing it. Although we can improve the quality of summary for a paper with its citation information, it cannot help authors to draft the summary while writing paper. 
Different to citation summarization, we generate the summary of the source paper by utilizing its reference papers as background knowledge. In our setting, the papers citing the source paper are not visible during the process of writing a summary.

\section{Semantic Scholar Network (SSN) Dataset}

Many scientific summarization datasets have emerged in recent years. The most commonly used scientific datasets, arXiv and PubMed \cite{cohan2018discourse}, focus on long document summarization without providing citation relationships between papers, which undoubtedly ignores the characteristics of the academic domain. 
\citet{yasunaga2019scisummnet} proposes a relatively small dataset containing 1k papers based on The ACL Anthology Network (ANN) \cite{radev2013acl}, but they generate summaries using only papers that cite the current paper (i.e., citing papers), which is unreasonable. In view of the above, we construct a large-scale summarization dataset, Semantic Scholar Network (SSN), consists of 141k research papers extracted from Semantic Scholar Open Research Corpus (S2ORC) \cite{lo2020s2orc}. All the papers in SSN form a large connected citation graph, allowing us to make full use of citation relationships between papers. 

\subsubsection{Dataset Preprocessing}
Semantic Scholar Open Research Corpus \cite{lo2020s2orc} contains 81.1M academic papers from multiple research fields. We only extract papers with full text \LaTeX{} parses (1.5M) which provides us more details about the paper (e.g. section names, boundaries of paragraph/sections). We keep papers whose abstract length is between 50 and 1000, and the body length is between 1000 and 8000.
Additionally, papers with less than 4 sections or do not have an Introduction section are also filtered out because they are likely to lose the discourse structure.
A Breadth-first search algorithm is applied to get a large connected citation graph. To prevent the graph from being too sparse, we recursively remove papers with only one 1-hop neighbour. We also normalize inline formulas, equations, tables, figures and citation markers with special tokens.

\subsubsection{Statistics}
SSN has 140,799 nodes and 660,908 edges where most papers come from the fields of mathematical, physics and computer science. Statistics of our dataset and other general datasets are shown in Table
~\ref{tab:statistics}. CNN/DailyMail \cite{hermann2015teaching} is a widely used news dataset, others belong to the scientific field. SSN has the longest text, which brings difficulty to modeling. Meanwhile, it has the most sections, showing that our dataset retains the most complete paper structure possible.
Besides, SSN is a connected huge citation graph,  indicating that SSN can be used to train some auxiliary tasks such as node classification and link prediction to help models better understand the research community in the whole graph.


\section{Method}

In this section, we first define the task of scientific papers summarization with citation graph, then describe our citation graph-based summarization model (\textsc{CgSum}) in detail.

\subsection{Problem Formalization}
Existing document summarization methods usually conceptualize this task as a sequence-to-sequence problem. Given a dataset $ D = (d_{1}, d_{2}, \ldots, d_{k})$, each document $d_i$ can be represented as a sequence of $n$ words $d$ $ = ({x}_{1}, {x}_{2},\ldots,{x}_{n})$, the objective is to generate a target summary $Y$ $ = ({y}_{1}, {y}_{2},\ldots,{y}_{m})$ by modeling the conditional distribution $p(y_{1},y_{2},\ldots,y_m|x_1,\ldots,x_n)$.

However, scientific papers have their own characteristics: there are citation relationships between papers, and the content of these papers is logically closely related. Therefore, we introduce the concept of citation graph to strengthen summarization tasks in the scientific domain. We define a citation graph $G = (V, E)$ on the whole dataset, which contains scientific papers and citation relationships. Each node $v \in V$ represents a scientific paper in the dataset, and each edge $e \in E$ indicates the citation relationship between two papers. Notably, when generating the summary of a paper, we cannot rely on the information of the papers that cites this one (because they are later in chronological order), so we extract a subgraph $G_v$ for each node $v$ to avoid introducing information that should not be used, the specific method can be seen in Algorithm \ref{algorithm:G_v}. 

\begin{algorithm}
		\caption{Citation Graph Construction}
		\begin{algorithmic}[1]
			\label{algorithm:G_v}
			\REQUIRE Node $v$; Citation graph of the whole dataset $G$
			\ENSURE Citation graph $G_v$ related to $v$
			\STATE Initialize a Queue $q$ and $G_v$ with Node $v$
			\WHILE{$q$ is not $\varnothing$}
			\STATE Dequeue Node $u$ from front of $q$
			\FOR{each Node $w \in G$ cited by $u$}
			\IF{$w$ $\not\in G_v$}
			\STATE Enqueue Node $w$ onto $q$
			\STATE Add Node $w$ to $G_v$
			\ENDIF
			\STATE Add Edge that $u$ cites $w$ to $G_v$
			\ENDFOR
			\ENDWHILE
		\end{algorithmic}
		\label{algorithm:G_v}
\end{algorithm}

Given the source paper $v$ (w/o abstract) and the related citation graph $G_v$ (we only use the abstract of other nodes), we need to generate a summary $Y$ of $v$ by modeling the conditional distribution $p(y_{1},y_{2},\ldots,y_m|x_1,\ldots,x_n;G_v)$.

\begin{figure}[t]
\centering
\includegraphics[width=0.5\textwidth]{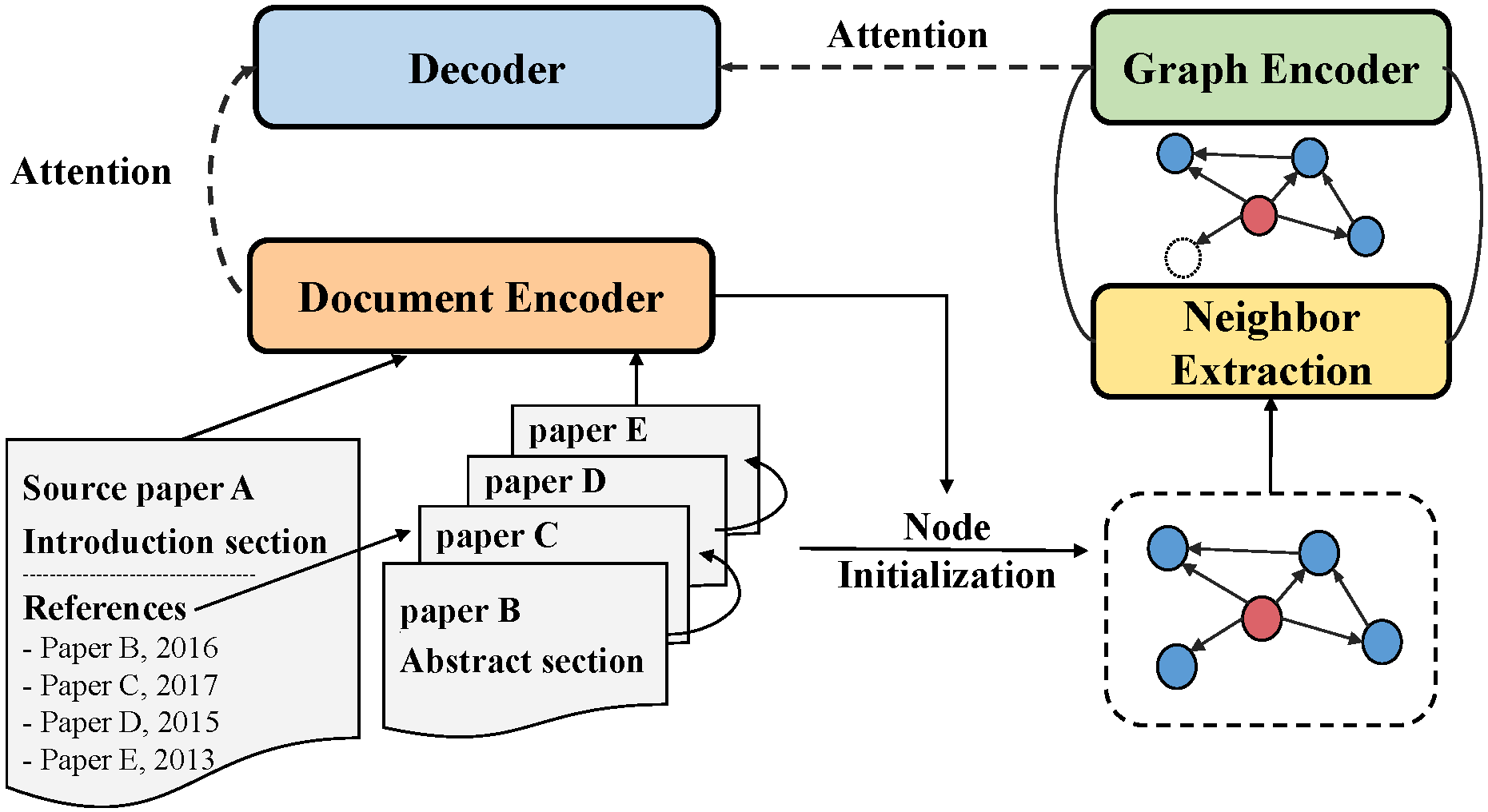}
\caption{
Overview of our Citation Graph-Based Model (\textsc{CgSum}).  A denotes the source paper (w/o abstract). B, C, D and E denote the reference papers. The body text of A and the abstract of reference papers are fed into the document encoder, and then used to initialize the node features in the graph encoder. Neighbor extraction method will be used to extract a more relevant subgraph. While decoding, the decoder will pay attention to both the document and the citation graph structure.
}
\label{fig:model}
\end{figure}
\subsection{Citation Graph-Based Summarization Model}

In this part, we illustrate our citation graph-based model (\textsc{CgSum}) as displayed in Figure \ref{fig:model}. The key idea is not only to encode the source document $v$, but also to capture the features of the corresponding citation graph $G_v$ to help us generate the summary. Our model consists of\textit{ document encoder}, \textit{graph encoder} and \textit{decoder}. In addition, we introduce a novel ROUGE credit approach.

\subsubsection{Document Encoder}
We employ a single-layer bidirectional LSTM (BiLSTM) to convert the input document $\bf d$ $ = ({x}_{1}, {x}_{2},\ldots,{x}_{n})$ to a sequence of hidden representations $\mathbf{H} = {\rm BiLSTM}({x}_{1},\ldots,{x}_{n})$. We initialize the source node $v_i$ by  pooling its hidden representations $\bf H$. For the neighbor nodes $v_j \in \mathcal{N} (v_i)$, where $\mathcal{N} (v_i)$ denotes the input neighborhood of $v_i$, we feed their abstract $t$ to another BiLSTM and obtain the initial representation $\mathbf{h}_{v_j}$  of node $v_j$ by aggregating the hidden representations of $t$ with a pooling layer.

\subsubsection{Neighborhood Extraction}

For each node $v$, it is too computationally expensive to use the whole citation graph $G_v$, so it is necessary to sample an informative subgraph. Specifically,
we first extract a directed subgraph $G_{v_i}'$ consisting of the source paper $v_i$ and its $K$-hop neighbors, and add self-loops to $G_{v_i}'$ for information enhancement. Before feeding $G_{v_i}'$ to the graph encoder,

we employ a neighborhood extraction method to further extract $T$ neighbors by their salience scores with source node $v_i$:

\begin{flalign}
\notag 
  s_{i,j} &={\rm softmax}(f([\mathbf{h}_{v_i}; \mathbf{h}_{v_j}]))\\
    &=\frac{exp(f([\mathbf{h}_{v_i}; \mathbf{h}_{v_j}]))}{\sum_{v_k \in G_v' }f([\mathbf{h}_{v_i}; \mathbf{h}_{v_k}])},
\end{flalign}
where $v_j \in G_{{v_i}}'$, $s_{i,j}$ denotes the salience score between $v_i$ and $v_j$, and $f$ is a 3-layers feed forward neural network.  We extract the most salient $T$ vertices with $\operatorname{argmax}$ function to construct the final citation graph $G^*_{v_i}$. However, directly sampling important nodes corrupts the training of parameters in $f$. To overcome this problem, we follow \citet{huang2020knowledge} and view $f$ as an information gate and multiplies $s_{i,j}$ to $v_j$ itself, $\mathbf{h}_{v_j} = s_{i,j}\mathbf{h}_{v_j}$.

\subsubsection{Graph Encoder}
Given a sampled citation graph $G^*_v$ and the initial nodes features $\mathbf{H}_v$, we use 2-layers graph attention networks (GAT) \cite{velivckovic2017graph} to update the representation of each node. Besides, to avoid the gradient vanishing problem, we add residual
connections between layers. $v_i$ is represented by the aggregation of its neighbors:

\begin{equation}
\label{eq:gat1}
\mathbf{h}_{v_i}^{'} = \mathbf{h}_{v_i} + \mathbin\Vert_{n=1}^{N} \sum_{v_j \in \mathcal{N} (v_i)} \alpha_{i,j}^n \mathbf{W}_v^{n} \mathbf{h}_{v_j},
\end{equation}

\begin{equation}
\label{eq:gat2}
\alpha_{i,j}^n =  {\rm softmax}(\mathbf{W}_a^n[
\mathbf{W}_q^{n} \mathbf{h}_{v_i};
\mathbf{W}_k^{n} \mathbf{h}_{v_j}]),
\end{equation}
where $\mathbin\Vert_{n=1}^{N}$ denotes concatenation of $N$ attention heads, and $\alpha_{i,j}^n$ is the normalized attention weight between $h_{v_i}$ and ${h_{v_j}}$ computed by the $n$-th attention head, $\mathbf{W}_a^n, \mathbf{W}_k^n, \mathbf{W}_q^n,\mathbf{W}_v^n$ are trainable parameters. Dropout \cite{hinton2012improving} with probability 0.1 is applied in each layer.

\subsubsection{Decoder}
Our decoder is a single-layer unidirectional LSTM. At each step $t$,  the decoder has a hidden state $\mathbf{s}_t$. Previous works \cite{see2017get} employ an attention mechanism to compute the attention distribution over the source words in the sequence-to-sequence structure, and we extend it to the graph structure as:

\begin{equation}
\label{eq:attn4}
e_{i,t}^{v}=\mathbf{v}^T {\rm tanh}(\mathbf{W}_h^v\mathbf{h}_{v_i}+\mathbf{W}_s^v\mathbf{s}_t+\mathbf{b}^v),
\end{equation}
\begin{equation}
\label{eq:attn5}
a_t^v = {\rm softmax}(\mathbf{e}_t^v),
\end{equation}
\begin{equation}
\label{eq:attn6}
\mathbf{h}_t^{v,*} = \sum_i a_{i,t}^v\mathbf{h}_i^{v},
\end{equation}
where $\mathbf{v}^T$, $\mathbf{W}_h^v$, $\mathbf{W}_s^v$ and $\mathbf{b}^v$ are trainable weights. We compute the attention distribution over the nodes in $G^*_v$ and obtain a graph context vector $\mathbf{h}_t^{v,*}$. Furthermore, on the basis of introducing the features of the citation graph, we still need to pay attention to the source document as:
\begin{equation}
\label{eq:mdattn1}
e_{i,t}=\mathbf{v}^T {\rm tanh}(\mathbf{W}_h\mathbf{h}_i+\mathbf{W}_s\mathbf{s}_t+ \mathbf{W}_v\mathbf{h}_t^{v,*}+\mathbf{b}),
\end{equation}
\begin{equation}
\label{eq:attn2}
a_t = {\rm softmax}(\mathbf{e}_t),
\end{equation}
\begin{equation}
\label{eq:attn3}
\mathbf{h}_t^* = \sum_i a_{i,t}\mathbf{h}_i,
\end{equation}
where $\mathbf{W}_h$, $\mathbf{W}_s$ and $\mathbf{b}$ are trainable parameters. $\mathbf{h}_t^{v,*}$ and $\mathbf{h}_t^*$ can be viewed as the aggregated representation of the citation graph and the source document respectively, so we concatenate them with the decoder hidden state $\mathbf{s}_t$ to produce the vocabulary distribution $P_{vocab}$:
\begin{equation}
\label{eq:p_vocab}
P_{vocab}={\rm softmax}(\mathbf{W}_o(\mathbf{W}_p[\mathbf{h}_t^{v,*}; \mathbf{h}_t^*;\mathbf{s}_t] + \mathbf{b}_o).
\end{equation}
In addition, to overcome the OOV problem, we allow the decoder to copy words from the source document as proposed by \citet{see2017get}. The
generation probability $p_{gen} \in [0,1]$ (i.e. the copying probability $p_{copy} = 1-P_{gen} $) for step $t$ is calculated as:
\begin{equation}
\label{eq:p_gen}
q_{gen}= \sigma(\mathbf{W}_c[\mathbf{h}_t^{v,*};\mathbf{h}_t^*;\mathbf{s}_t;\mathbf{x}_t] + \mathbf{b}_c),
\end{equation}
where $\mathbf{x}_t$ denotes the decoder input at time step $t$. Therefore, the probability distribution over the extended vocabulary is:

\begin{equation}
P_{final} = q_{gen}P_{vocab} + (1 - q_{gen})\sum_{i:w_i=w}a_{i,t}.
\end{equation}
Obviously, if $w$ does not appear in the source document, $\sum_{i:w_i=w}a_{i,t}$ is equal to zero, and if $w$ is an OOV word, $P_{vocab}$ is zero. The loss at time step $t$ is the negative log likelihood of the target word $y_t$: 
\begin{equation}
\label{eq:log_loss}
 \mathrm{loss}_t = -\mathrm{log}(y_t|v;\theta),
\end{equation}
where $v$ is the source document and $\theta$ are the parameters of our model. We add an coverage loss to penalize repeatedly attending to the same word in the source document. $\mathrm{covloss}_t =\sum_i \mathrm{min}(a_{i,t} ;c_{i,t}),$ where $ c_{i,t} =\sum_{t^{'}=0}^{t-1}a_{i,{t^{'}}}.$ Finally the  overall loss for the whole sequence is:
\begin{equation}
    \mathrm{loss} = \frac{1}{T}\sum_{t=0}^T(\mathrm{loss}_t + \lambda*\mathrm{covloss}_t),
\end{equation}
where $\lambda$ is the hyperparameter to reweight the coverage loss.

\subsubsection{ROUGE Credit}
Intuitively, the information brought by the citation graph is not only useful during training, but it is also helpful for the model to generate summaries during inference. Motivated by this, we propose a novel ROUGE credit score in beam search algorithm to instruct our model to write summaries with the help of nearby nodes' abstracts.

Specifically, at the decoding step $t$, we first select the neighbor $nbr_{max}$ which has the most influence on the generated summary using $\operatorname{argmax}$ function over the attention distribution $a_t^v$ on the graph $G^*_v$. In the beam search process, there are $k$ candidate sequences $C = (c_1, \ldots, c_k)$ per time step, then we calcaute the ROUGE credit score between the abstract of $nbr_{max}$ and $c_i$ as:

\begin{equation}
\label{eq:rougeCredit}
  credit_{i} = {\rm ROUGE}(\mathrm{abst}[nbr_{max}], c_{i})*g(t)
\end{equation}

\begin{equation}
g(t)=
\begin{cases}
1 & t < l_s\\
exp(1- l_s/t) & t \geq l_s
\end{cases}
\end{equation}
where $\mathrm{abst}[nbr_{max}]$ represents the abstract of the selected neighbor, $g(t)$ is a weight function corresponding to the decoding step  $t$ and $l_s$ is a hyperparameter (if $t \geq l_s$ the credit score will take more weight). We design $g(t)$ by simply modifying the sentence brevity penalty function in BLEU \cite{papineni2002bleu}, which makes the final generated summary neither bias towards the abstract of neighbor nodes, nor focus on the words selected by the model on the vocabulary. 
At last, the total score of the $i$-th candidate summary $c_i$ is given by the sum of its average log likelihood and $credit_{i}$. In our experiments, we calculate this credit every 5 steps as a trade-off to decoding time.

\section{Experiments}

\subsection{Dataset}
\begin{figure}[t]
\centering
\includegraphics[width=0.9\linewidth]{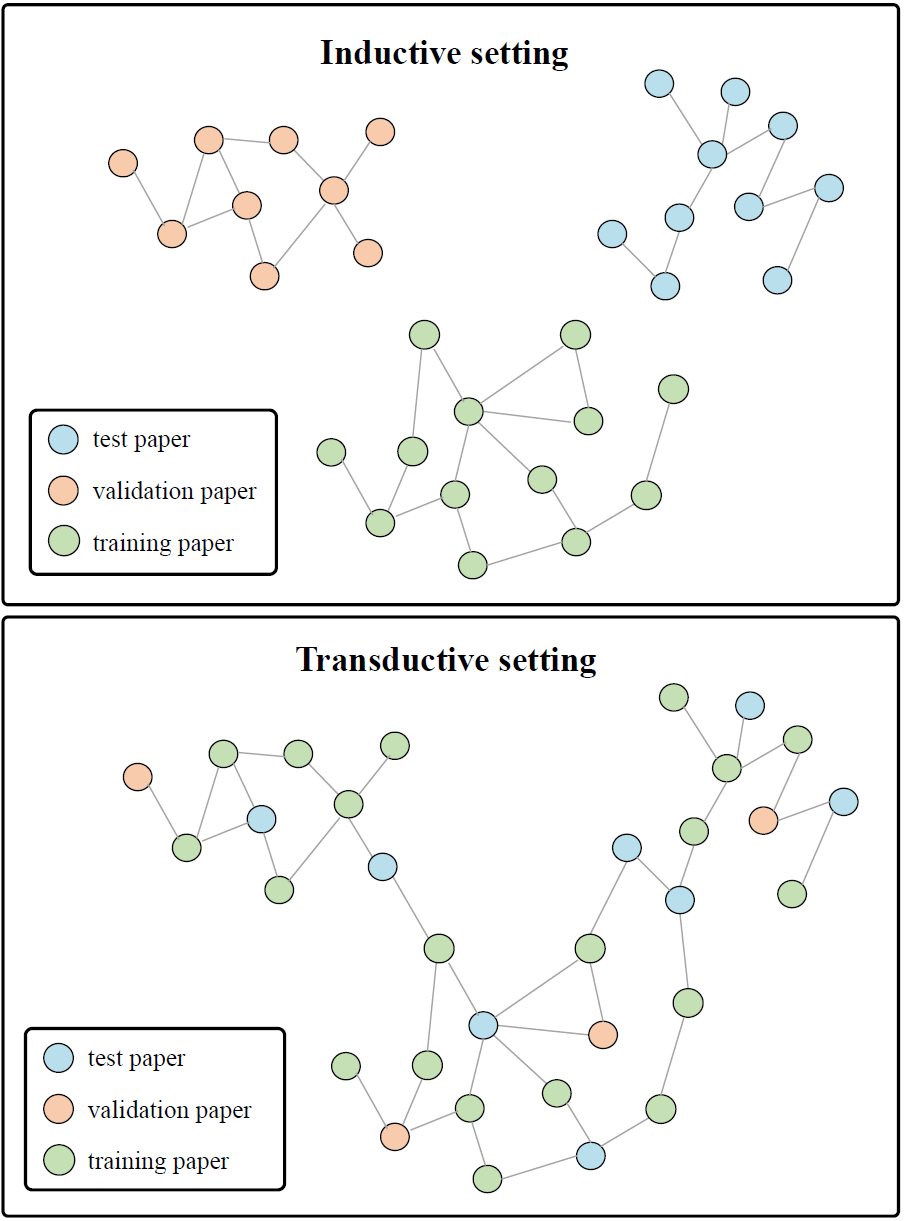}
\caption{
Different ways of splitting training, validation, test sets from the whole graph. We omit the directionality of the edges for simplification. The green, orange, cyan nodes represent papers from the training, validation, test set.
}
\label{fig:split}
\end{figure}

We evaluate our model on our Semantic Scholar Network dataset. Details about our dataset is shown in Table \ref{tab:statistics}. We lowercase all
tokens and tokenize sentence and word using spaCy \cite{honnibal-johnson-2015-improved}.
As is shown in Figure \ref{fig:split}, for SNN (transductive) we randomly choose 6,250/6250 papers from the whole dataset as test/validation sets and the remaining 128,299 papers are classified as training set which is the most commonly way to split the dataset.
The transductive division indicates that most neighbors of papers in test set are from the training set, but considering that in real cases, the test papers may from a new graph which has nothing to do with papers we used for training, thus we introduce SNN (inductive), by splitting the whole citation graph into three independent subgraphs -- training, validation and test graphs with the breadth first search algorithm. The training/validation/test graphs in inductive setting contain 128,400/6,123/6,276 nodes and 603,737/17,221/14,515 edges. In both inductive and transducitve setting, the summary of papers in the test set and validation set are kept invisible during the training phase. Our inductive setting also has the intention to test whether models trained in a large-scale citation graph has the ability to transfer to another citation graph. Therefore, the inductive setting of our task is thought more difficult.

\subsection{Training Details and Parameters}
We use hyperparameters suggested by \citet{see2017get} in the BiLSTM model. The word embeddings layer is trained from scratch without  using any pretrained language models and embeddings.
We use mini-batches of size 16 and we limit the input document length to 500 tokens. The input citation graph includes the source paper and its $K$-hop neighbors ($K$ = 1, 2), and we initialize the node representation with body text of source papers and the abstract of neighbors. We constrain the maximum number of papers in an input graph to 64. We implement graph attention network with Deep Graph Library \cite{wang2019dgl} and the number of attention heads is set to 4. We use Adagrad optimizer with learning rate 0.15 and an initial accumulator value of 0.1. We set the beam size $b$ to 5 and $l_s$ to 75 in the ROUGE credit, and the ROUGE \cite{lin2004rouge} score used is the value of ROUGE-1 $\rm F_1$. We do not train the model with coverage loss in the first epoch to help the model converge faster, and we train our model for 10 epochs and do validation every 2000 steps. We select the best checkpoint based on the ROUGE-L score on the validation set.

\subsection{Baseline Methods}
We provide the \textsc{LEAD} baseline  which extracts the first $L$ (depending on the number of sentences in the reference summary) sentences from the source document and \textsc{Oracle} as an upper bound of extractive summarization systems. We use a greedy algorithm following \citet{nallapati2016summarunner} to generate an oracle summary. Since we truncate the document to 500 tokens,\textsc{Oracle} in this paper is calculated on the truncated datasets.

Besides, we implement the following extractive systems: (1)  \textsc{TextRank} \cite{mihalcea2004textrank}: an unsupervised extractive system based on the graph structure (2) Transformer\textsc{Ext}: an extractive system based on transformer encoders (3) \textsc{BertSumExt} \cite{liu2019text}: an extractive summarization model with BERT.
We further add the following abstractive baseline models:
(1) \textsc{PTGen}+\textsc{Cov} \cite{see2017get}: an abstractive summarization system with copy mechanism. (2) Transformer\textsc{Abs}: an abstractive summarization model based on transformer (3) \textsc{BertSumAbs}\cite{liu2019text}: an abstractive summarization system built on BERT.
We employ trigram blocking \cite{paulus2017deep} to reduce redundancy for both the baseline systems and our models.

\subsection{Experimental Results}
\begin{table}[!tb]
\tabcolsep0.03 in
\begin{tabular}{lccc}
\toprule
\textbf{Systems} \vrule width 0pt height 11.5pt depth 5pt & \textbf{R-1} & \textbf{R-2} & \textbf{R-L} \\
  \midrule
  \textsc{Oracle}\vrule width 0pt depth 0pt &\textcolor{white}{{\bf *}}51.04\textcolor{white}{{\bf *}} & 23.34 & \textcolor{white}{{\bf *}}45.88 \textcolor{white}{{\bf *}}\\
  \textsc{LEAD}\vrule width 0pt height 0pt  & 28.29 & 5.99 & 24.84 \\
 \midrule
 \hspace{-1.5mm} {\textbf{ Extractive }} \vrule width 0pt height 12pt depth 5pt &   &  &    \\
 
\textsc{TextRank} \vrule width 0pt depth 0pt & 36.36 & 9.67 & 32.72 \\
Transformer\textsc{Ext}\vrule width 0pt height 0pt  &  43.14 & 13.68 & 38.65 \\
\textsc{BertSumExt} \vrule width 0pt height 0pt depth 5pt & 42.41& 13.10 & 37.97 \\
\textsc{BertSumExt} (\textit{mp} = 640) \vrule width 0pt height 0pt depth 5pt & 44.28 & 14.67 & \textbf{39.77} \\

\midrule

\hspace{-0mm}{\textbf{Abstractive}} \vrule width 0pt height 12pt depth 5pt &   &  &   \\
\textsc{PTGen} + \textsc{Cov}\vrule width 0pt height 0pt  & 42.84 & 13.28 & 37.59 \\
\; \textit{Concat Nbr. Summ} \vrule width 0pt height 0pt  & 43.05 & 13.53 & 37.97 \\
Transformer\textsc{Abs}\vrule width 0pt height 0pt  & 37.78 & 9.59 & 34.21\\
Transformer\textsc{Abs} + \textsc{Copy}\vrule width 0pt height 0pt  & 43.35 & 14.87 & 39.17 \\
\textsc{BertSumAbs}  \vrule width 0pt height 0pt depth 5pt & 41.22& 13.31 & 37.22 \\
\textsc{BertSumAbs} (\textit{mp} = 640) \vrule width 0pt height 0pt depth 5pt & 43.73 & \textbf{15.05} & 39.46\\
\; \textit{Concat Nbr. Summ (mp=}640)  \vrule width 0pt height 0pt depth 5pt &43.45& 14.89 & 39.27 \\

\midrule
 
\hspace{-0.5mm}{\textbf{Our Model}} \vrule width 0pt height 12pt depth 5pt &   &  &   \\

 \textsc{CgSum} + 1-hop Nbr.\vrule width 0pt height 0pt  & \textbf{44.36} & 14.69 & 39.43 \\

\textsc{CgSum} + 2-hop Nbr.  \vrule width 0pt height 0pt  & 44.28 & 14.75 & \textbf{39.76} \\

\bottomrule
\end{tabular}
\caption{Results on SSN (inductive). \textit{Concat Nbr. Summ} denotes the input is a concatenation of source document and neighbors' summary, \textit{mp} means the expanded size of position embedding in BERT. \textsc{CgSum} denotes our Citation Graph-Based Summarization Model.}
\label{tab:results}
\end{table}

\begin{table}[!tb]
\tabcolsep0.05 in
\begin{tabular}{lccc}
\toprule
\textbf{Systems} \vrule width 0pt height 11.5pt depth 5pt & \textbf{R-1} & \textbf{R-2} & \textbf{R-L} \\
  \midrule
  \textsc{Oracle}\vrule width 0pt depth 0pt &\textcolor{white}{{\bf *}}50.12\textcolor{white}{{\bf *}} & 23.31 & \textcolor{white}{{\bf *}}45.29 \textcolor{white}{{\bf *}}\\
  \textsc{LEAD}\vrule width 0pt height 0pt  & 28.30 & 6.87 & 24.93 \\
 \midrule
 \hspace{-1.5mm} {\textbf{ Extractive }} \vrule width 0pt height 12pt depth 5pt &   &  &    \\
 
\textsc{TextRank} \vrule width 0pt depth 0pt & 40.81 & 12.81 & 36.47 \\
Transformer\textsc{Ext}\vrule width 0pt height 0pt  &  41.45 & 13.02 & 37.20 \\
\textsc{BertSumExt} \vrule width 0pt height 0pt depth 5pt & 41.68 & 13.31 & 37.42 \\

\textsc{BertSumExt} (\textit{mp} = 640) \vrule width 0pt height 0pt depth 5pt & 43.23 & 14.59 & 38.91 \\

\midrule

\hspace{-0mm}{\textbf{Abstractive}} \vrule width 0pt height 12pt depth 5pt &   &  &   \\

\textsc{PTGen} + \textsc{Cov}\vrule width 0pt height 0pt  & 39.46 & 12.06 & 35.72 \\
\; \textit{Concat Nbr. Summ} \vrule width 0pt height 0pt  & 40.12 & 12.58 & 35.94 \\
Transformer\textsc{Abs}\vrule width 0pt height 0pt & 36.58 & 10.19 & 33.13 \\
Transformer\textsc{Abs} + \textsc{Copy}\vrule width 0pt height 0pt  & 40.83 & 14.71 & 36.93 \\

\textsc{BertSumAbs}  \vrule width 0pt height 0pt depth 5pt & 40.38 & 14.07 & 36.54 \\

\textsc{BertSumAbs} (\textit{mp} = 640) \vrule width 0pt height 0pt depth 5pt & 41.92 & \textbf{15.09} & 37.79\\

\; \textit{Concat Nbr. Summ (mp=}640)  \vrule width 0pt height 0pt depth 5pt &41.11& 14.50 & 37.16 \\

\midrule
 
\hspace{-0.5mm}{\textbf{Our Model}} \vrule width 0pt height 12pt depth 5pt &   &  &   \\
 \textsc{CgSum} + 1-hop Nbr.\vrule width 0pt height 0pt  & 43.10 & 14.90 & \textbf{39.10} \\
\textsc{CgSum} + 2-hop Nbr.  \vrule width 0pt height 0pt  & \textbf{43.45} & 14.71 & 38.89 \\

\bottomrule
\end{tabular}
\caption{Results on SSN (transductive).}
\label{tab:results_trans}
\end{table}

\begin{figure}[t]
\centering
\includegraphics[width=1\linewidth]{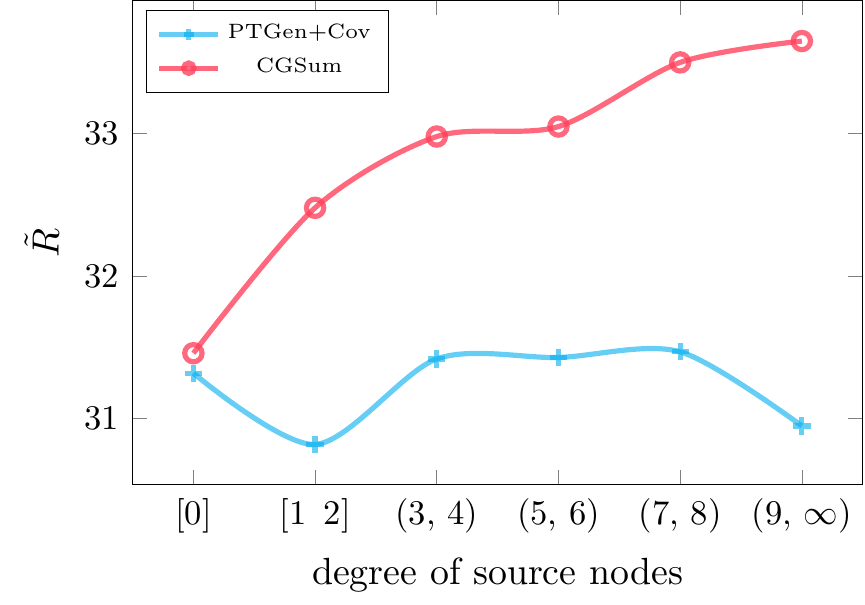}
\caption{
Relationships between the  degree of
source paper nodes (X-axis) and $\Tilde{R}$ (the average of ROUGE-1, ROUGE-2 and ROUGE-L)  of two models: \textsc{CgSum} + 1-hop neighbors 
and \textsc{PTGen} + \textsc{Cov} (inductive setting).
}
\label{fig:degree}
\end{figure}

\subsubsection{Reusult on SSN}
We evaluate summarization quality with the standard ROUGE score \cite{lin2004rouge} where R-1 and R-2 represent informativeness and R-L represents fluency.
Table~\ref{tab:results},\ref{tab:results_trans} show the results on our dataset. 
Several well-known extractive and abstractive baselines as well as  models that make use of pretrained language model BERT \cite{devlin2018bert} using their open-sourced implementations are shown in the second and third part. Besides, to better compare our model with the baseline models, for each abstractive baseline we give an additional \textit{Concat Nbr. Summ} version whose input is the concatenation of source document and the neighbors' abstracts separated by a special token \texttt{[SEP]} following the general setting in multi-documents summarization. 
In our experiments, we are surprised to find that Transformer\textsc{Abs} performs poorly on our dataset, but it will be significantly improved if we further add copy mechanism.

Although BERT has achieved the state-of-the-art performance in the News domain \cite{DBLP:conf/acl/ZhongLCWQH20}, it has not shown great advantages in the field of scientific papers. We think the main reason here is that  BERT has a length limit of 512, but scientific papers are usually much longer than this limit. To solve this issue, we break the constrain on maximum length in BERT by adding more position embeddings which are initialized randomly and finetune them in the training phrase, which brings remarkable improvement for two BERT-based models (\textsc{BertSumExt} and \textsc{BertSumAbs}).
In addition, all models have not significantly improved after adding the content of the cited papers (i.e., \textit{Concat Nbr. Summ}), which shows that the content of the reference papers is not enough.

As can be seen from Table \ref{tab:results},\ref{tab:results_trans}, in both inductive and transductive settings, \textsc{CGSum} outperforms all the pretrained models in terms of R-1 and R-L metrics. When compared with \textsc{BertSumAbs} (\textit{mp} = 640), which is also an abstractive model, although our model uses a shorter input sequence (500 vs 640) and a simpler encoder structure (1-layer BiLSTM and 2-layers GAT vs 12-layers pretrained transformers), it still outperforms \textsc{BertSumAbs} (\textit{mp} = 640). This result fully illustrates a combination of the document information and the features of the citation graph structure can greatly help the model better understand the relevant research community, thereby naturally generating high-quality abstracts. In inductive setting, \textsc{CGSum} beats BERT by 0.63 R-1 score and beats \textsc{PTGen} + \textsc{Cov} by 1.52  R-1 score  while  \textsc{CGSum} brings more significant improvements in transductive setting (beats BERT by  1.53 R-1 score and beats \textsc{PTGen} + \textsc{Cov}  3.99 R-1 score).

\subsubsection{Degree of Source Paper}
We further explore the relationship between model performance and the degree of the source node. We divide our test set into six parts according to the degree of the node. As is shown in Figure \ref{fig:degree}, there is no obvious connection between the performance of \textsc{PTGen} + \textsc{Cov} and the degree of the source paper. Notably, \textsc{PTGen} + \textsc{Cov} can be viewed as our model removes the graph encoder, so for the nodes with degree 0, the two models have similar performance. However, as the degree of nodes increases, our model can gradually achieve better performance. This dataset splitting experiment shows that our model is good at handling papers with rich citation graph information, that is to say, an informative and relevant research community is very important for understanding a scientific paper. 
In inductive setting the average degree of nodes $d_{avg} = 2.3 $ in test set is much smaller than that in transductive setting  $d_{avg} = 4.7$. This experiment also gives an explanation of why \textsc{CGSum} outperforms other baseline models without using citation graph by a larger margin in the transductive setting. 

\begin{table}[t]
    \centering
    \begin{tabular}{lccc}
        \toprule
        \textbf{Model}  & \textbf{R-1} & \textbf{R-2} & \textbf{R-L} \\
        \midrule
        \textsc{CgSum} & \textbf{44.36} & \textbf{14.69} & \textbf{39.43} \\
        \quad - Nbr. Extraction &44.23 & 14.63 &39.29\\
        \quad - Residual Connection &44.25 & 14.41 & 38.95\\
        \quad - Trigram Blocking &43.48 & 14.49 & 38.92\\
        \quad - ROUGE Credit & 43.81 & 14.49 & 38.70 \\
        \quad - GNN Encoder &42.84 & 13.28 & 37.59\\
        \bottomrule
    \end{tabular}
    \caption{Ablation study of the \textsc{CgSum}. ’-’ means we remove the module from
the original \textsc{CgSum} (inductive setting).} 
    \label{tab:Ablation}
\end{table}

\subsubsection{Ablation Study}
To have a better understanding of the contribution of each component in our proposed model, we remove the neighborhood extraction, residual connection, trigram blocking, rouge credit and GNN from the origin model. As shown in Table \ref{tab:Ablation}, neighborhood extraction obtains a certain performance improvement because it extracts a more informative subgraph. Residual connection and trigram blocking have been proven to work well in previous work, and they are also effective in our task. Besides, our proposed ROUGE credit method significantly improve the performance on R-1 and R-L because of the shared domain-specific terms and the similar writing style among papers in the same research community. Finally, if we remove the GNN encoder, our model actually become \textsc{PTGen} + \textsc{Cov}.

\section{Conclusion}
In this paper, we augment the task of scientific papers summarization with the citation graph.
Specifically, summarization systems can not only use the document information of the source paper, but also find the useful information from the corresponding research community from citation graph to generate the final abstract. 
Different to the previous work, we aim to help researchers draft a paper abstract by utilizing its references, rather than the papers citing it. We construct a large-scale scientific summarization dataset which is a huge connected citation graph with 141K nodes and 661K citation edges. We also design a novel citation graph-based model which incorporates the features of a paper and its references. Experiments show the effectiveness of our proposed model and the important role of citation graphs for scientific paper summarization.

\section{Acknowledgements}
We would like to thank the anonymous reviewers for their
valuable suggestions. This work was supported by the National Key Research and Development Program of China (No. 2017YFB1002104).
\bibliography{aaai21}
\bibliographystyle{aaai21}
\end{document}